%% file: draft.tex
\documentstyle[epsfig]{nemlap}
\begin{document}
\title{Isolated-Word Confusion Metrics and the PGPfone Alphabet}
\author{Patrick Juola}
\institute{Department of Experimental Psychology \\ Oxford University \\
South Parks Road \\ Oxford, UK  OX1 3UD \\ {\tt patrick.juola@psy.ox.ac.uk } }

\maketitle

\begin{abstract}
Although the confusion of individual phonemes and features have been studied
and analyzed since Miller and Nicely \cite{miller-nicely}, there has been
little work
done on extending this to a predictive theory of word-level confusions. 
The PGPfone alphabet is a good touchstone problem for developing
such word-level confusion metrics.  This paper presents some difficulties
incurred, along with their proposed solutions, in the extension of phonetic
confusion results to a theoretical whole-word phonetic distance metric.
The proposed solutions have been used, in conjunction with a set of
selection filters, in a genetic algorithm to 
automatically generate appropriate word lists for a radio alphabet.
This work illustrates some principles and pitfalls that should be
addressed in any numeric theory of isolated word perception.
\end{abstract}

\section{Motivations}
\begin{figure}[htbp]
  \begin{center}
    \leavevmode
    \psfig{figure=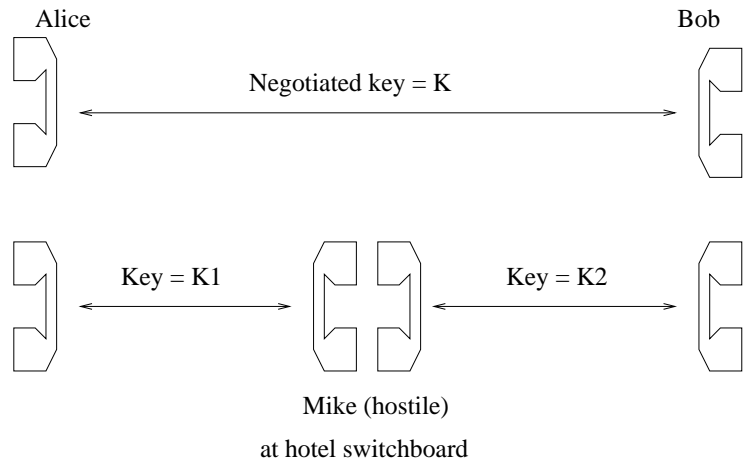,height=6.00cm}
  \end{center}
  \caption{\label{figure:manmiddle}Man-in-the-middle attack scenario}
\end{figure}

The PGPfone project \cite{pgpfone}, developed by Boulder Software
Engineering, provides high-quality secure voice communications over
ordinary phone lines.  Implicit in this project, as in any security
project, is the need to keep the (voice) data and the keys used for
scrambling/encrypting the data secure from eavesdroppers---or from
hostile listeners who may be able to do more than simply tap
telephones.  In normal operations, Alice can simply telephone
Bob using PGPfone.  The program will negotiate an encryption
``key'' between Alice
and Bob and transform the data so that any eavesdropper cannot understand
their conversation without the key.  Although this works against passive
eavesdropping, it is not secure against a more powerful adversary.  If
the hostile Mike, who works at the hotel where Alice is staying, can
arrange to intercept or reroute her calls, he can arrange that her call
goes to him, while he himself places the call to Bob.  Mike's computer
can then
negotiate two keys (one with Bob and one with Alice) while Mike
connects the two conversations to each other, and notes everything
said.
Figure~\ref{figure:manmiddle} illustrates this basic
``man-in-the-middle'' attack.
However, if
each call uses a separate encryption key (as in PGPfone), and if Alice
and Bob can confirm that they are using the same key, they can
be relatively confident that there is no Mike in the middle.

Reading long binary (or even hexadecimal) strings
over the phone is, however, tedious and error-prone.  There are many other
applications that might require this sort of data exchange.  For instance,
one standard method of confirming that a key you obtained is valid is by
calculating a ``key fingerprint'' and checking it against the owner.
From the owner's perspective, it is difficult to remember the string
of random characters that comprise the fingerprint, and from the reader's
perspective, it is imperative that the words be understood properly
at the other end of a telephone line.  Some other applications are
reading signatures or remembering keys---and in all cases, it would
be more efficient and accurate to use some sort of encoding for compression
and error-checking.  

\section{Alphabet Design}

PGPfone's designed solution to this problem is to develop a word
list, styled after
the traditional military or pilot's alphabet (alpha, bravo, charlie, \ldots),
with each word representing some
fixed number of bits.
The problem of developing such a linguistic encoding for data exchange
is unusual in that, unlike most NLP projects, ``language'' here
is one of the independent variables that can be manipulated at the
engineer's whim.  The words used in developing this alphabet, if
properly chosen, can not
only provide compression, but can  also provide some error
prevention, error detection, and a considerable human factors advantage.
On the other hand, ``proper chosing'' will in turn be helped by
an efficient, accurate, and
numerical model of the desired properties.

For example, the length of the word list, obviously, will determine some of its
attributes.  A small list (for example, sixteen words) would provide
no compression over reading hexadecimal numbers, but could still provide
some degree of errorproofing by removing potentially confusing tuples
like five/nine, B/C/D/E,
and so forth.  A list of 64 words would allow about 30\% compression (over
hexadecimal digits)
in terms of number of words, but the complete list would be much harder
for a human to memorize. 
For situations where humans are required to generate
keys (and/or responses) from memory, this would be an unreasonable expectation.
However, in PGPfone, all keys are generated and stored by the computers,
and the only job for a human is to read a series of words presented by
the computer; thus, there is no need for a human to ever memorize
the complete list.
Because these keys are going to be generated automatically from a
list known only to
a computer, we opted to use lists of 256 words, allowing each word to
represent a byte.   Using larger lists would obviously allow significantly
better compression, but require considerably more (computer) memory
to store the word table.  For example, two lists of 256 words can be
stored in only 5 kilobytes of memory.  A larger list (two bytes
per word) would require nearly 650 kilobytes of memory, as well as
a word vocabulary larger than most native English speakers' productive
vocabulary.

From a human factors perspective,
an ideal word list consists of short,
easily recognizable and easily pronounceable words with easily distinguished
prefixes and a minimum of phonetic confusibility or bad associations.
We chose to approach this task as a selection problem---from a much
larger list,
select words with appropriately chosen
characteristics.
For this project, we used the Moby Pronunciator database, distributed
by Grady Ward,
which contains nearly 200,000 word/pronunciation pairs.
In some characteristics, such as ``short'', the selection process is
a trivial task.  In others, such as ``no bad associations'', this is
nearly impossible to perform automatically and it was recognized that this
would need to be done by hand.  The main technical difficulty that
we considered to be solvable by computer occurred in the representation
of ``phonetic confusibility.''

The ideal metric for phonetic confusibility for this project would
be capable of accepting any two speaker-independent representations of
word pronunciations and returning, as a distance, an accurate measure
of the probability
of one word being confused for the other in an isolated word context.
In practical terms, this is probably an unachievable goal. For example,
part of the mathematical definition of distance includes the notion of
symmetry, that if one word has a fixed probability of confusion with
the second word, the second word has an identical probability of
being confused with the first.  One may naively expect this to fail.

Even a vaguely correct metric could have far wider applications than the
simple PGPfone alphabet, however.  For instance, Nakisa and Hahn
\cite{nakisa-hahn} describe a model for the German plural system based
on the notion of mapping novel words to appropriate inflectional
categories based on the phonetic properties of the word.  In one experiment,
they simply calculated the ``nearest neighbor'' using a Euclidean distance
of a 240-element feature representation.  Clearly, the more accurate
the distance representation used, the more confidence one can have in
their (psycholinguistic) conclusions. Furthermore, an accurate
confusibility measure could have important engineering applications, for
example to the development of case-based reasoning tools for
text to speech system or speech synthesis.  Finally, because mathematical
tools like this permit language and speech to be the object of manipulation
instead of mere study, this has wider applications in any
controlled-language situation, for example the development of simplified
language for MT projects or the production of distinctive brand
or product names.

\section{Linguistic distances}
Our approach to the problem of phonetic confusibility is a variant 
of the work of Miller and Nicely \cite{miller-nicely}.  In particular, words are
ordered strings of phonemes instead of acoustic signals, phonemes
in turn contain features [such as those enumerated in 
Ladefoged \cite{ladefoged}], and individual
phonemes can be meaningfully compared by comparing their features.
It is further assumed that 
the phonetic distance between
two words can be 
approximated by some function of the differences between
the phonemes that comprise the words.

It should be noted immediately that this is only one of
many possible approaches.  For example, Lindblom \cite{lindblom} measured
vowel similarity
based on formant frequency, and in particular the F1 and F2
frequencies.  However, this approach is less satisfactory for several
reasons :  sounds, especially consonants, show
much more word-to-word variance than
phonetic features; although the phonetic transcription of a given
word does not often vary from person to person, the actual acoustic
signal does; and thirdly, the simple task of mapping from sound to lexeme
is itself a hard problem, while taking little if anything away from
the difficulties in designing a distance metric.  Other approaches have
been proposed using ``autosegmental phonology'' as suggested by
Goldsmith \cite{goldsmith} to compress word representations into feature change
sets, at the expense of synchronization data.  Strictly speaking,
Miller and Nicely \cite{miller-nicely} and more recently
Bell\footnote{Alan Bell, 1995. Personal communication.} use a more
introspective/scientific approach than simply calculating 
mathematical distances, instead directly examining people's
perceived distances, which may or may not exactly map onto a feature-based
metric---but this approach requires either extensive lab-work to validate,
or a willingness to rely on pure introspection without validation.

Unfortunately, the chosen approach almost immediately encounters severe
representational difficulties at a number of levels.  For instance,
phonologists and phoneticians usually use
feature representations designed to represent differences
important to the production of a sound.  The amount of detail, and
hence importance, thus changes with the degree of variability in
a feature.   Sounds with voiceless stop
consonants can be produced at many different locations ranging from the
lips to the very back of the throat.  Voicing, on the
other hand, is either present
or absent; no known language makes a distinction between voiceless,
strongly-voiced, and weakly-but-still-voiced consonants.  However,
Miller and Nicely \cite{miller-nicely} indicate that voicing is one
of the most salient and robust features of English consonants; in other
words that /d/ is
more likely to be misheard as /g/ than as /t/ (under most circumstances).
Generalizing this, we have the unfortunate result that phoneme pairs may
differ in several unimportant features and yet sound closer than another
pair that differ only in one extremely salient aspect.

Furthermore, the relative salience of features varies wildly depending upon
the sort of noise in which the signal is embedded.  Given that 
system designers have no idea of the conditions under which people may
use a telephone, the best one can reasonably do is to make assumptions; in
this case, we assume white noise.

Although the Miller-Nicely confusion matrices provide exact data that
could be used to balance some features, they don't provide enough data
for our purposes.  The
study only incorporated differences between some English consonants and
no vowel distinctions at all.  Because of the absence of such data,
standard automatic feature weighting or pattern recognition
techniques seemed inapplicable;
instead, we relied on the balance information from Miller-Nicely,
applied as best we could to the entire featural universe.
Ladefoged \cite{ladefoged} proposes a more extensive
list of features that allow for all sounds of English, consonants and
vowels alike, to be presented and distinguished on a uniform scale.  As
discussed above, this list provides no data on salience, but with 
appropriate judgements (and 
some coercion of scales between vowels and consonants), the various
features can be
approximately balanced to the Miller-Nicely data.
Using this method, the perceptual
difference between two comparable phonemes can be measured as the number of
bits that differ in the two representations. 

The final representation developed for the PGPfone alphabet is attached as
table~\ref{table:features}.  Multivalued features, such as place, were
represented as ``thermometer codes''.  Binary features, such as voicing,
were of course merely on or off, replicated enough times to achieve the
desired weight.  The final representation requires 26 bits per phoneme.

\begin{table}
\centering
\begin{tabular}{|lll|}
\hline
Feature name & Sample & Number of bits \\
\hline \hline
Place of articulation & /d/ vs /g/ & 7 \\
Manner of articulation & /l/ vs /t/ & 6 \\
Height of articulation & /i/ vs /$\epsilon$/ & 5 \\
Voicing & /z/ vs /s/ & 4 \\
Syllabic & vowels vs. cons. & 1 \\
Nasal & /n/ vs /d/ & 1 \\
Lateral & /l/ vs /r/ & 1 \\
Roundedness & (various) & 1 \\
Sibilant & /s/ vs /f/ & not used \\
\hline
\end{tabular}
\caption{\label{table:features}Phoneme coding for PGPfone alphabet}
\end{table}

Even granting the viability and success of a phoneme-by-phoneme
perceptual distance metric, there are difficulties in its extension
to full-word distances, and here theory provides less support than
might be wished.  For example, if each phoneme were weighted equally
and could be directly compared with a single other phoneme, the
difference between two words can be as simple as the sum of the 
phoneme differences.  However, some phonemes are clearly more
salient than others.  On a gross level, the stressed syllables of
a word pair are intuitively of much greater salience than the unstressed
ones.  Furthermore, psycholinguistic results like
Slobin \cite{slobin} or Derwing and Neary \cite{derwing-nearey}
suggest that onsets
are more salient than codas.  These
results unfortunately provide little suggestion about whether a simple
weighting will
address this disparity, or what weights would be most appropriate.
The approach taken in PGPfone was a simple one; the preceding consonant cluster
and vowel(s) of the stressed syllable were given twice normal weight,
as was the (word-)onset phoneme.

A similar problem arises with non-aligned or non-existent sounds.
For example, should the word /bEst/ be treated as most similar to
/bEts/, /bEt/, or /bEs/?  Derwing and Neary \cite{derwing-nearey}
present a few primitive
metrics to address this question, based on primarily on a  notion
of sequences of identical vs. nonidentical phonemes.  A more sophisticated
approach could use the notion of ``edit-distance'' as typified by Myers
\cite{myers},
but only with an accurate measure of the perceived difference between a
sound and its absence, or in other words, a featural representation for
silence.  The representation of silence has produced some interesting
opinions [for example, Cottrell and Plunkett \cite{cottrell-plunkett}
presented silence as a
voiced, nasal, sibilant, back vowel], but little agreement or theory.

This problem can be reduced by the use of templates.  For instance, if all
the words in
a study are of the form CVC, then there need be no
representation of silence as all phonemes are aligned directly.  If the
words can be coerced into such a form, for example by elimination of
words with consonant clusters, then fewer sound/silence comparisons are
necessary.  For
independent reasons (discussed below), the PGPfone list demands
words with small consonant clusters and of a particular syllabic
structure.  The list to be selected from was filtered before the
selection process began to eliminate unsuitable words with long consonant
clusters.  By increasing
the strength of the prefiltering, one
can restrict attention to words where comparisons are meaningful, or
phrased another way, one can greatly limit the damage done by a bad
representation of silence.

Similarly, by careful use of duplicate sounds, some of the silence/sound
comparisons can be avoided.  Vowel sounds can be lengthened
or shortened almost at will, thus, vowel blends (such as /Oi/)
are compared with ``pure'' vowels such as /i/ by the simple technique
of presenting the pure vowel twice (/ii/) and comparing---and thus
/Oi/ is accurately represented as mid-way
between /O/ (/OO/) and /i/ (/ii/).  Similar
tricks could be used to tease apart different consonant clusters; for
example, fricative (but not stop) consonants could be extended as
vowels above, or some simple combinatorics might apply to compare all
possible alignments and select one.

One final concern for the PGPfone distance metric touches on the
incorporation of additional, non-linguistic features.  For example,
it would be nice if the final words had distinguishable orthographic prefixes,
to make it easier for keyboard entry and similar (non-linguistic)
processing tasks.
Obviously, paying attention to such things will, in theory, negatively
impact the linguistic quality of the final solution but result in
a better system overall.  For the PGPfone list, orthographic spread
was achieved by appending to each phonological representation an
ASCII representation of the first two {\em characters} of the word.

Once the representation is in place, the actual distance was calculated
as the number of bits that differ between two word representations.
The word level distance metric (for the two-syllable template)
is attached as table~\ref{table:wordcoding}.  The three syllable template,
of course, is similar except for the additional consonant and vowel, 
and a slight increase in complexity in the representation of the
stressed syllable.

\begin{table}
\centering
\begin{tabular}{|ll|}
\hline
Phonetic aspect &  Number of bits \\
\hline \hline
Onset consonant(s) & 78 \\
First syllable vowel  & 52 \\
Middle consonant(s) & 52 \\
Second syllable vowel & 52 \\
Final consonant(s) & 52 \\
Initial characters & 12 \\
Stressed vowel & 52 \\
Stress pattern & 7 \\
\hline
\end{tabular}
\caption{\label{table:wordcoding}Word coding for PGPfone alphabet}
\end{table}

\section{Engineering Aspects}

The ultimate test of any representation is the quality of solutions
it permits.  A good solution for the PGPfone list involves several
additional qualities than simply an accurate distance representation,
as detailed in this section.

Humans, when reading sequences, tend to make different errors than simple
bit-flips (misreadings), so error detection and recovery is a bit different
than simply correcting bits.  Instead, humans tend to
either omit, duplicate, or switch (adjacent) words, rather than misread
them.  Furthermore, humans tend not to be able to do complex Boolean
arithmetic in their heads, and so full error correction is usually not
practical.  Stewart\footnote{Zhahai Stewart. 1991. Personal communication
to Philip Zimmermann} suggested a clever way to
allow human-like errors to be easily detected.  By building {\em two} lists
instead of one, and alternating the lists from which the words in the
sequence come, one can easily spot any such errors by noticing that
two successive words come from the same list.  This assumes, of course,
that the (listening) human can tell from which list a word has been drawn.
The two lists for PGPfone are obviously different in that one consists only
of two syllable words, and the other of three.

Similarly, the lists should consist of words that are easily pronounceable
and easily readable.  Words with multiple spellings or multiple pronunciations
(including cases like Polish, the nationality, vs. polish, the
cleaning product) are perilous because they may be read or transcribed
incorrectly---and accordingly were deleted from the lists without
consideration.  Similarly, any words for which we had evidence of
significant phonetic variations (e.g. tomato) were removed.  Furthermore,
as there might be a significant pool of list users with difficulty with
some sounds or clusters, any hard-to-pronounce words, defined
as words incorporating any non-English sounds or lengthy consonant
or vowel clusters, were also eliminated.

Unfortunately, the filtering methods chosen do not readily solve the
problem of dialect or language variance.  As an obvious example,
the words in the Moby Pronunciator database are given with their
pronunciations {\em in an American, and specifically Pacific,
accent.}  Although the California accent is relatively neutral within
the United States, it's certainly not neutral or standard worldwide.
Even within the United States, dialect differences can make even as
fundamental questions such as the number of syllables problematic.
Words containing semivowels such as ``oilcan'' and ``fragile''
have been suggested
as being three syllable words in some dialects of (American) English,
and as two syllable words in others, although the actual phonetic
data on this is unclear.  Further afield, native language
differences can also raise difficulties.  Japanese and Chinese, for
example, are notorious for not distinguishing the ``lateral'' feature
between /l/ and /r/; the phoneme weighting of this feature has been
artificially reduced, but not eliminated, in an effort to balance its
relevance to the English speaking community and its irrelevance to
(parts of) the Pacific rim.  To attempt to solve this in the filtering
process, for example by eliminating all words with semivowels or
lateral consonants, would have resulted in a list of candidate words
too small to be useful.

The most difficult aspect of the list to control was unfortunately
one of the more important; the final lists should contain words with
appropriate associations.  One of the goals was to develop a word list
that would inspire a certain amount of confidence in the security of
the overall product.  The standard military/pilots' alphabet, for example,
has a certain ``coolness,''  the same mystique that applies to a
child's Captain Midnight secret decoder rings.  An ideal list would
capture the same indefinable feeling.  And although it proved difficult
enough to banish repugnant words (for instance, the computer
selected ``nigger'' from the dictionary in an early test), there seems
no automated procedure for detecting all and only ``cool'' words.

We were forced to rely on what ad hoc principles we could identify.  The
standard pilots' alphabet, for instance, contains familiar but
uncommon words; several words do not appear in the Brown corpus at all,
while no word listed has a frequency of 85 occurrences or more.
So for the PGPfone list, words
that were too unfamiliar or too common were eliminated.
The filtering to get appropriate consonant clusters
seemed to help here as well.  Empirically, nouns seemed better
than verbs, which in turn seemed better than adjectives, but all three were
substantially better than the rest of English, but appropriate 
databases were not available to automate and make use of that
observation.  In general,
noninflected words seem stronger than their inflected variants.
In the end, we
were forced to rely on human judgement, generating a list, blue-pencilling
or modifying
words that we found inappropriate, then using the survivors as the base
for another list.

Once the selection and measuring criteria are available, the actual selection
of the list is, technologically speaking, near-trivial.  Because of
the high dimensionality of the search space, direct solution of the
best subset in the candidate was held to be infeasible.  (Such algorithms
tend to be either polynomial in the dimension of space to be searched, or
exponential in the number of elements in the candidate list.)
Instead, we opted to
use a standard multivariate approximation technique to find an
acceptable partial solution that could be refined as needed.  Several
algorithms could be used for this; for example, tabu search, a
recent variation on hill climbing with momentum, was briefly considered
but also rejected due to the high dimensionality.  Simulated annealing
is another standard optimization technique, but lacks a strong enough
element of incremental learning.  As the initial stage in simulated
annealing is typically the ``melting'' of the entire knowledge base,
any useful knowledge from an initial approximation will be entirely
lost on the second and subsequent attempts at a solution.

For these reasons, we used a simple
genetic algorithm \cite{GA} to evolve a
near-optimum (sub)set of the candidates
such that the smallest distance between any pair was maximized.
Genetic algorithms (GAs) have been widely used as a general-purpose black-box
optimization algorithm, and their use here has no wider implications
beyond simply being a known, uncontroversial, and effective method
of solving optimization problems.  Specifically,
the GA generated a population of random 256-word
subsets of the
candidate list.  Subsets were permitted to ``breed'' by trading some of
their members, and the daughter subsets were evaluated (using the distance
metric described above) to determine the
closest pairwise distance.  Successful children were allowed to 
be fruitful and multiply, while the losers in the genetic sweepstakes were
simply dropped from the population.  After several hundred generations,
the top candidate was then edited as described above, and the surviving words
were used as
a fixed and unchanging part of the entire population for the next run of
the GA selection program.  

It proved necessary as well to cross-check the list pairs.  For example,
the word ``guitar'' is phonetically distinct in English (being one of
the few words where a hard g preceeds a short i).  Unfortunately, the
word ``guitarist'' is phonetically distinct for the same reason.  Because
the comparison scheme used was template based, there was no easy way
to automatically compare words from the two lists and calculate a
numeric distance.  Instead words from one list which were derivationally
related to words from the other list were individually inspected, and
usually the less ``cool''  element of the relational pairs (most often
the base or uninflected form) was hand-eliminated.

After several runs, when a final, accepted list had been agreed upon,
the words in each list were alphabetized without regard to case
and used to represent byte values from 0 to 255.  Some sample words from
the middle of the lists are here attached as table~\ref{table:words}.

\begin{table}
\centering
\begin{tabular}{|lll|lll|} \hline 
Number & 2 syllable & 3 syllable & Number & 2 syllable & 3 syllable \\
\hline \hline
111 & glucose & hesitate & 116 & guidance & impartial \\
112 & goggles & hideaway & 117 & hamlet & impetus \\
113 & goldfish & holiness & 118 & highchair & inception \\
114 & granny & hurricane & 119 & hockey & indigo \\
115 & gremlin & hydraulic & 120 & hotdog & inertia \\
\hline
\end{tabular}
\caption{\label{table:words}Sample words from the PGPfone list}
\end{table}

Figure~\ref{figure:inuse} shows an example of the list in use in a
nonPGPfone context.  The large block of nearly opaque text is
a cryptographic key for a program called ``PGP.''  Using this key,
anyone can send secret mail to the author.  The block of hexadecimal
digits, the ``key fingerprint'' can be used to quickly confirm that the
key has been received properly.  It can easily be seen that the same
function can be performed more accurately, quickly, and memorably by
the encoded fingerprint.

\input{minipage}

\section{Implications and Conclusions}

The final alphabet as distributed in PGPfone appears to work well enough
for the purpose for which it was designed; our feedback has generally been
positive, and suggested improvements tend to be matters of opinion on
single words rather than major changes to the underlying structure or
model.  This work does strongly suggest the need for further work on
the development of word-scale phonological confusibility models.  The
alphabet itself might have been made much stronger if we had been able to take
several dozen subjects into a phonetics laboratory and test the weightings
we conjectured above.  Fundamental data on the salience of various
word-level characteristics is available only in a very sketchy manner
(and likely to vary significantly with language anyway.)

Clearly, a full evaluation of this work requires some empirical
checking, which at this point has not yet been done.  Although informal
tests show that the words are understood, the degree of confusibility
has not been rigorously tested.  There are many open questions that
are grounds for future work.  How confusible {\em are} the words?
Does the actual transmission channel correspond to the assumptions
used in the feature weights?  Do the assumptions of a reasonable,
obvious, and unique pronunciation fail when the reader is not a native
English speaker?

Although this problem may seem artificial in many regards, it lends itself
well to treatment as a touchstone problem for many speech/language
generation problems.  The difficulty we encountered with the representation
of consonant clusters mimics the difficulties other researchers such as
Cottrell and Plunkett \cite{cottrell-plunkett} or MacWhinney \cite{macwhinney}
have had with the learning
and representation
of sound patterns in language acquisition tasks.  Particularly in situations
such as neural networks or supervised learning, where a distance measure
is used to direct the system to its new state, an accurate distance measure
is more a necessity than a convenience.  An accurate statistical analysis
of the effectiveness and salience of various feature-based representations
may shed light to bridge the sound/phoneme gap---as well as help with the
(word) segmentation problem and provide fundamental evidence about the
psychological reality of phonemes and phonetic features.

From an engineering perspective, an accurate way of measuring perceptual
distance could help in any situation where language must be engineered to
fit a particular need.  This could be of use, for instance, in sublanguage
selection and generation, or more prosaically to help with the creation
of novel and distinctive product and service names. 

This work illustrates several basic principles that a reasonable metric
should follow: 
\begin{itemize}
\item  First, that standard feature sets
do not accurately reflect the perceived salience of
various features.
\item  Second, that feature differences are a significant but not 
all-encompassing
part of the perceived differences among words; superphonemic attributes
such as stress and onset must also be taken into account.
\item  Third, that templates are best used to control the sorts of
comparisons and measurements taken, but that using them will greatly
restrict the overall validity of the measurements.
\end{itemize}
There are almost certainly other principles that could be found and
added to this list.  It is hoped that future work, whether in the
context of PGPfone 2.0 or other unrelated projects, will be able to
extend this list of principles to a full theory of isolated word perception.


\end{document}

%% file: minipage.tex
\begin{figure*}
\centering
\begin{minipage}{4.5in}
\begin{verbatim}
-----BEGIN PGP PUBLIC KEY BLOCK-----
Version: 2.6.2i

mQCNAzDIPcEAAAEEALvWEowkZJ8sLUnOcMkCykWpjKirlwEv3LAC6c6ciU63bhzn
yVcH22KKZQj6n+A2sIn+qLdKiKlLNOd0Bh7wIwlJrlYb/g6zMyw6TpWPPRopzkis
7U2eofSKZ4L19RSVw8+QejFvHeMx89+QdTzUNXTAAthkJZporyC+v3X+p5ZhAAUR
tCpQYXRyaWNrIEp1b2xhIDxwYXRyaWNrLmp1b2xhQHBzeS5veC5hYy51az6JAJUD
BRAw5a+WZXmEuMepZt0BAfSdA/0VQcMu1oV8N1Tx4MTI8gk/FN7BYH5PHFpF0QrQ
Ahr4NZKN395q7LvMPb6jbsuLAI2eamg6ujQZU3X5iiXMS58dm7F7ATz0PRVh9768
dl62STyMNVMBbYc2Wqruk7jDHIw2HaU+8CMSWJE66FbKO8y7TnAy1TTIXOvHR6OL
EOWLbw==
=uB9I
-----END PGP PUBLIC KEY BLOCK-----

Key fingerprint =  5C 39 76 D5 DD E2 9E C2  56 2C A2 91 C7 91 65 F9
Encoded fingerprint =
escape crossover hotdog speculate swelter torpedo puppy reproduce
egghead combustion quota molecule spaniel molecule fracture Waterloo
\end{verbatim}
\end{minipage}
\caption{\label{figure:inuse}Author's PGP key, with encoded fingerprint}
\end{figure*}